\documentclass{article} 
\usepackage{style/iclr_2026/iclr2026_conference,times}

\usepackage[utf8]{inputenc} 
\usepackage[T1]{fontenc}    
\usepackage{hyperref}       
\usepackage{url}            
\usepackage{booktabs}       
\usepackage{amsfonts}       
\usepackage{nicefrac}       
\usepackage{microtype}      

\usepackage{amsmath,amsfonts,bm}

\def\eqref#1{equation~\ref{#1}}

\def\1{\bm{1}}

\def\rvx{{\mathbf{x}}}

\def\vx{{\bm{x}}}

\DeclareMathAlphabet{\mathsfit}{\encodingdefault}{\sfdefault}{m}{sl}
\SetMathAlphabet{\mathsfit}{bold}{\encodingdefault}{\sfdefault}{bx}{n}

\usepackage{multirow}
\usepackage{tablefootnote}
\usepackage{makecell}
\usepackage{enumitem}
\usepackage{soul}
\usepackage[export]{adjustbox}
\usepackage{bbm}
\usepackage{bbding}
\usepackage{minitoc}
\usepackage{graphicx}
\usepackage[table, svgnames]{xcolor}
\usepackage{subfig}
\usepackage{cleveref}
\usepackage{tikz}
\usepackage{mathtools}
\usepackage{tcolorbox}
\usepackage{lscape}
\usepackage{wrapfig}

\usepackage{amsmath}
\usepackage{amssymb}
\usepackage{mathtools}
\usepackage{amsthm}

\usepackage{algorithm} 
\usepackage{algorithmic}  
\usepackage[algo2e]{algorithm2e} 

\newcommand\independent{\protect\mathpalette{\protect\independenT}{\perp}}
\def\independenT#1#2{\mathrel{\rlap{$#1#2$}\mkern2mu{#1#2}}}

\newcommand*\circled[1]{\tikz[baseline=(char.base)]{\node[shape=circle,draw,inner sep=0.5pt] (char) {#1};}}

\theoremstyle{plain}

\theoremstyle{definition}

\theoremstyle{remark}

\newcommand{\titlecontent}{TabStruct: Measuring Structural Fidelity of Tabular Data}
\title{TabStruct: Measuring Structural Fidelity\\of Tabular Data}

\author{
  Xiangjian Jiang\textsuperscript{1},\space Nikola Simidjievski\textsuperscript{2,1},\space Mateja Jamnik\textsuperscript{1} \\
  \textsuperscript{1}Department of Computer Science and Technology, University of Cambridge, UK \\
  \textsuperscript{2}Télécom Paris, Institut Polytechnique de Paris, France\\ 
  {\scriptsize \texttt{xj265@cam.ac.uk, nikola.simidjievski@telecom-paris.fr, mateja.jamnik@cl.cam.ac.uk}} \\
}

\iclrfinalcopy 
\begin{document}
\doparttoc
\faketableofcontents

\maketitle

\begin{abstract}
\looseness-1
Evaluating tabular generators remains a challenging problem, as the unique causal structural prior of heterogeneous tabular data does not lend itself to intuitive human inspection. Recent work has introduced structural fidelity as a tabular-specific evaluation dimension to assess whether synthetic data complies with the causal structures of real data. However, existing benchmarks often neglect the interplay between structural fidelity and conventional evaluation dimensions, thus failing to provide a holistic understanding of model performance. Moreover, they are typically limited to toy datasets, as quantifying existing structural fidelity metrics requires access to ground-truth causal structures, which are rarely available for real-world datasets.
In this paper, we propose a novel evaluation framework that jointly considers structural fidelity and conventional evaluation dimensions. We introduce a new evaluation metric, \textit{global utility}, which enables the assessment of structural fidelity even in the absence of ground-truth causal structures. In addition, we present \textit{TabStruct}, a comprehensive evaluation benchmark offering large-scale quantitative analysis on 13 tabular generators from nine distinct categories, across 29 datasets. Our results demonstrate that global utility provides a task-independent, domain-agnostic lens for tabular generator performance. We release the TabStruct benchmark suite, including all datasets, evaluation pipelines, and raw results.
Code is available at \url{https://github.com/SilenceX12138/TabStruct}.
\end{abstract}

\begin{figure}[!t]
    \centering
    \includegraphics[width=0.88\textwidth]{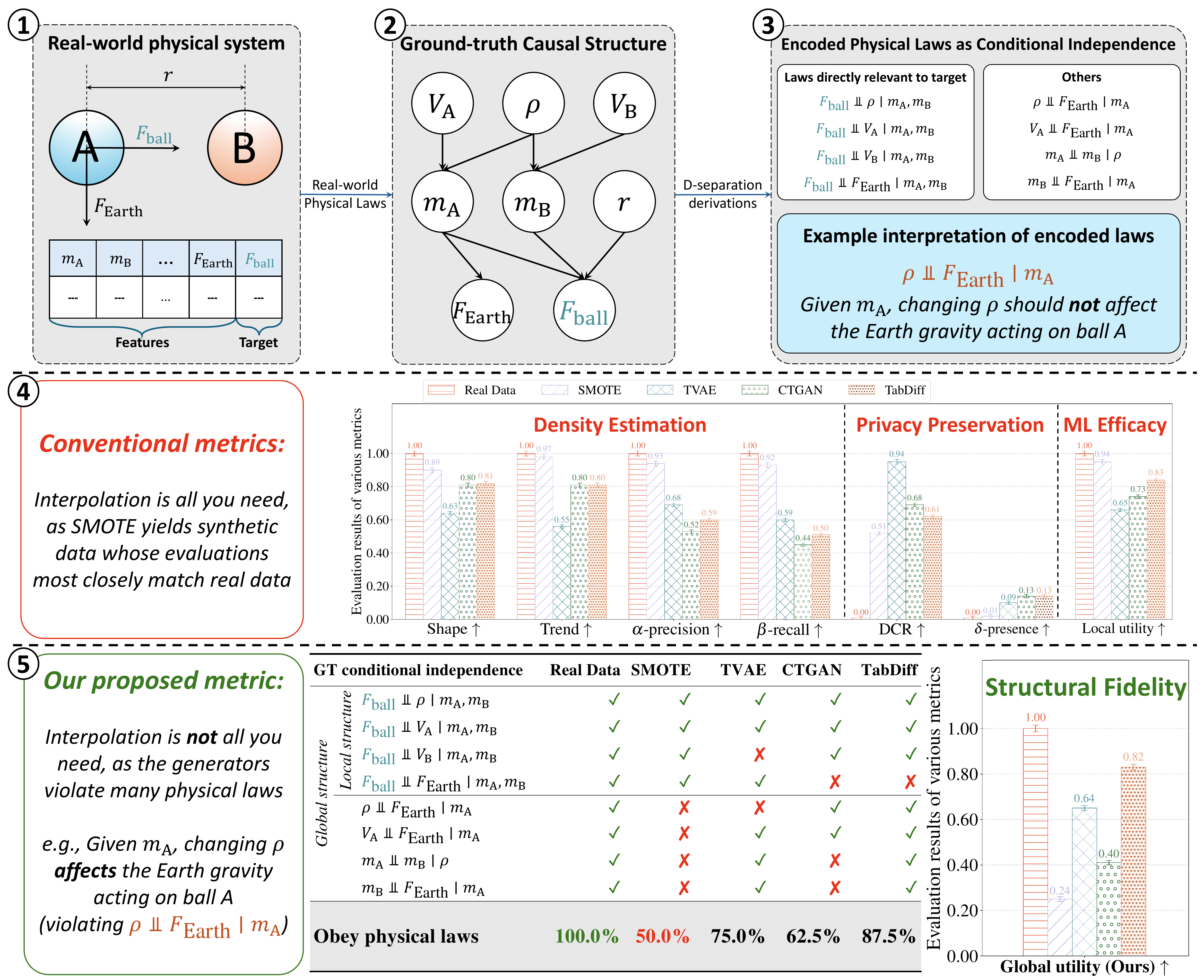}
    \caption{\textbf{Illustrative example highlighting the importance of fidelity check for tabular data structure.} \textbf{\protect\circled{1}:}~A real-world physical system showing the gravitational forces acting on ball A. The system is described by ball density ($\rho$), volume ($V$), masses ($m_{\text{A}}$ \& $m_{\text{B}}$), distance ($r$), and gravitational forces ({\color[HTML]{008080} $F_{\text{ball}}$} \& $F_{\text{Earth}}$). For simplicity, we assume both balls share identical density. \textbf{\protect\circled{2}:}~We derive the ground-truth (GT) causal structure of the system based on Newton's law of universal gravitation. \textbf{\protect\circled{3}:}~We interpret the encoded physical laws of the system as the conditional independence (CI) across variables. \textbf{\protect\circled{4}:}~We evaluate four generators by conventional metrics. \textbf{\protect\circled{5}:}~We assess the structural fidelity by CI tests and the proposed global utility metric. We note that the \textit{global structure} reflects full conditional independence across all variables, while the local structure includes only those directly relevant to a specific prediction task at hand ({\color[HTML]{008080} $F_{\text{ball}}$}). Results demonstrate that conventional metrics are insufficient:  for instance, while SMOTE is able to outperform other generators on conventionally used dimensions (e.g., ML efficacy) -- the generated synthetic data only preserves local structure and violates most physical laws. For tabular data, where the truthfulness and authenticity of synthetic data is hard to verify, global utility provides an effective mechanism for evaluating the alignment of the synthetic data to the likely ground-truth causal structure.}
\label{fig:motivation}
\vspace{-5mm}
\end{figure}

\section{Introduction}
\label{sec:intro}

\looseness-1
Tabular data generation is a cornerstone of many real-world machine learning tasks~\citep{borisov2022deep, fang2024large}, ranging from training data augmentation~\citep{margeloiutabebm, cui2024tabular} to missing data imputation~\citep{zhangmixed, shi2024tabdiff}. 
These applications underscore the importance of generative modelling, which necessitates an appropriate understanding of the underlying data structure~\citep{kingma2014auto, goodfellow2014generative, bilodeau2022generative}. For instance, textual data conforms to the distributional hypothesis, and thus the autoregressive models are a natural workhorse for the text generation process~\citep{zhao2023survey, sahlgren2008distributional}. 
In contrast to the homogeneous modalities like text, tabular data can pose a different structural prior due to its heterogeneity -- the features within a dataset typically have varying types and semantics, with feature sets that can differ across datasets~\citep{grinsztajn2022tree, shi2024tabdiff}. Recent work~\citep{hollmann2025accurate} on tabular foundation predictors has empirically demonstrated that the Structural Causal Model (SCM) is an effective structural prior of tabular data. 
As such, it is important to investigate how effectively existing tabular generative models capture and leverage the causal structures. 

\looseness-1
Prior work~\citep{hansen2023reimagining, qian2024synthcity, du2024systematic, tu2024causality, livieris2024evaluation, kapar2025what} has attempted to assess tabular data generators by evaluating the synthetic data they produce. However, the prevailing evaluation paradigms still exhibit three primary limitations, which are summarised in~\Cref{tab:literature-benchmark-overview}:
\textit{(i)~Insufficient tabular-specific fidelity assessments.}
Current benchmarks largely adopt evaluation dimensions from homogeneous data modalities, such as density estimation~\citep{alaa2022faithful}, machine learning (ML) efficacy~\citep{xu2019modeling}, and privacy preservation~\citep{kotelnikov2023tabddpm}. While effective in other modalities, they exhibit conceptual limitations when applied to tabular data -- they do not explicitly assess the unique structural prior of tabular data. A notable example is that many generators (e.g., SMOTE) can produce synthetic data with similar density estimation as real data, yet still violate underlying causal structures -- such as physical laws illustrated in~\Cref{fig:motivation}(\circled{3}). Although CauTabBench~\citep{tu2024causality} takes a step forward to assess the structural fidelity of synthetic data, it remains confined to toy SCM datasets (i.e., synthetic datasets derived from random SCMs), offering limited insight into real-world tabular data, where the ground-truth SCMs are unavailable.
\textit{(ii)~Potential evaluation biases.}
Many benchmarks~\citep{hansen2023reimagining, qian2024synthcity} and model studies~\citep{xu2019modeling, margeloiutabebm, zhangmixed} prioritise ML efficacy as the principal dimension for assessing generator performance. For instance, in a classification setting, a generator is often considered effective if its synthetic data allows downstream models to achieve high predictive performance. However, while useful, ML efficacy can be highly sensitive to the choice of prediction task and target (\Cref{fig:motivation}(\circled{5}) and \Cref{sec:CI_scores}). 
\textit{(iii)~Limited evaluation scope.}
Existing benchmarks mainly consider only a narrow range of datasets and generative models (\Cref{tab:literature-benchmark-overview}), which restricts their ability to provide a thorough and generalisable comparison of model performance across the broader landscape of tabular generative modelling.

\begin{figure}[!t]
    \centering
    \includegraphics[width=\linewidth]{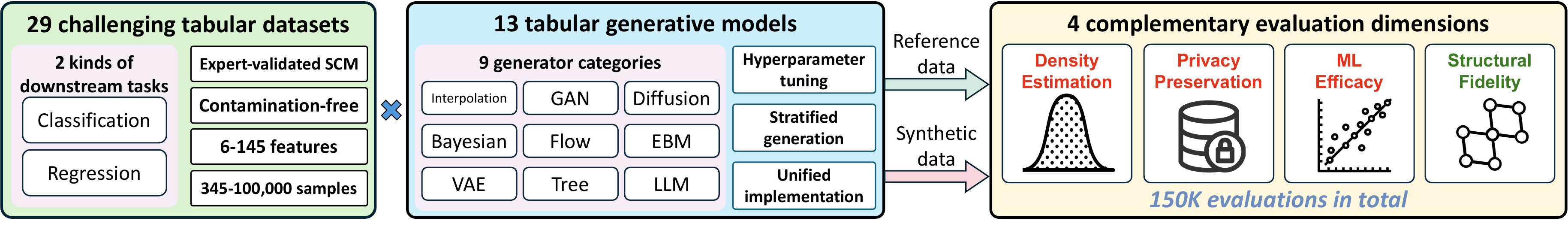}
    \caption{\textbf{Overview of the proposed evaluation framework.} TabStruct provides a comprehensive evaluation benchmark, including structural fidelity and conventional dimensions, for 13 representative tabular generative models on 29 challenging tabular datasets.}
    \label{fig:framework}
\end{figure}

In this paper, we aim to bridge these gaps by introducing a systematic and comprehensive evaluation framework for existing tabular generative models, with a particular focus on the structural prior of tabular data. Our proposed framework is characterised by five key concepts:
(i)~We explicitly incorporate \textit{structural fidelity} of synthetic data as a core evaluation dimension for tabular generative models. Structural fidelity can directly reflect model capability in learning the structure of tabular data, without biasing towards a specific prediction target. In addition, we investigate its interplay with three conventional evaluation dimensions, offering customised guidance for selecting suitable generators across diverse use cases.
(ii)~We evaluate structural fidelity on expert-validated SCM datasets. To ensure alignment with ground-truth causal structures, we avoid using toy SCMs and instead select SCM datasets with expert-validated causal structures. With ground-truth SCMs, we can quantify structural fidelity through the difference in \textit{conditional independence (CI)} between real and synthetic data as shown in~\Cref{fig:motivation}(\circled{5}).
(iii)~We further extend the evaluation of structural fidelity to real-world datasets, where the ground-truth SCMs are unavailable. To this end, we propose a novel evaluation metric, \textit{global utility}, which treats each variable as a prediction target and measures how well it can be predicted using other variables. Importantly, global utility does not require ground-truth causal structures, thus enabling the evaluation of structural fidelity in real-world scenarios.
(iv)~We conduct an extensive empirical study on the performance of \textit{13 tabular generators spanning nine categories on 29 datasets}, resulting in a total of \textit{over 150,000 evaluations}. The large evaluation scope can ensure holistic and robust benchmarking results.
(v)~We introduce \textit{TabStruct} (\Cref{fig:framework}), the benchmark suite developed for this work. This open-source library aims to help the research community explore tabular generative modelling within a standardised framework.

Across both SCM and real-world datasets, our primary finding is: 
\vspace{-2mm}
\begin{center}
    \textit{Structural fidelity, as quantified by the proposed global utility, should be a core dimension when evaluating tabular generative models.}
\end{center}
\vspace{-2mm}
The benchmark results suggest the prevailing paradigm (i.e., optimising tabular generators primarily for improved density estimation and ML efficacy) is insufficient. In contrast, our proposed global utility provides insights into a crucial yet underexplored perspective -- tabular-specific fidelity assessments.
Our contributions can be summarised as follows:

\begin{itemize}[topsep=0pt, leftmargin=10pt, itemsep=0pt]
    \item \textbf{Conceptual}~(\Cref{sec:method}): We propose a unified evaluation framework for tabular generators that integrates structural fidelity with conventional dimensions, and introduce \textit{global utility}, a novel metric that measures structural fidelity without requiring access to ground-truth causal structures.
    \item \textbf{Technical}~(\Cref{sec:method}): We release the \textit{TabStruct} benchmark suite, including datasets, generator implementations, evaluation pipelines, and all raw results.
    \item \textbf{Empirical}~(\Cref{sec:exp}): We conduct a large-scale quantitative study of 13 tabular generators on 29 datasets. The results offer actionable insights into model performance and can inspire the design of more effective tabular generators by attending to the unique structural prior of tabular data.

\end{itemize}

\section{Related Work}
\label{sec:literature_review}

\looseness-1
\textbf{Tabular Generator Benchmarks.}
An extensive line of benchmarks~\citep{stoian2025survey, hansen2023reimagining, qian2024synthcity, du2024systematic, kindji2024under, sidorenko2025benchmarking, long2025evaluating} has been proposed for tabular data generation, conventionally established around three dimensions: density estimation, privacy preservation, and ML efficacy. 
Mainstream evaluation metrics typically capture specific aspects of inter-feature interactions. However, they rarely assess whether the underlying causal structures are preserved in the generated tabular data.

Density estimation~\citep{hansen2023reimagining, alaa2022faithful, shi2024tabdiff, zhangmixed} assesses the divergence between real and synthetic data distributions. However, it fails to explicitly capture inter-feature causal interactions. 
ML efficacy~\citep{xu2019modeling, qian2024synthcity, seedat2024curated, tiwald2025tabularargn} evaluates the performance difference when real data is replaced with synthetic data in downstream tasks, which primarily focuses on $p(y \mid \rvx)$, thus inherently prioritising feature-target relationships over inter-feature interactions.
Privacy preservation~\citep{du2024systematic, kotelnikov2023tabddpm, hu2024sok, espinosa2023quality}, although essential in privacy-sensitive scenarios, is generally task-specific and usually does not necessitate high structural fidelity~\citep{chundawat2022universal, livieris2024evaluation, mclachlan2018aten}.
Recent efforts such as Synthcity~\citep{qian2024synthcity} and SynMeter~\citep{du2024systematic} have aimed to standardise the evaluation of tabular data generators by incorporating the three conventional dimensions. Nonetheless, they omit explicit assessment of tabular data structure. 
To the best of our knowledge, CauTabBench~\citep{tu2024causality} is the only other benchmark to explicitly evaluate structural fidelity, but it is limited to toy SCM datasets, as existing metrics~\citep{chen2023structured, spirtes2001causation} typically assume access to the ground-truth SCMs -- a condition that is seldom satisfied and arguably infeasible for most real-world datasets~\citep{kaddour2022causal, glymour2019review, zhou2024ocdb}. 
In addition, some prior studies~\citep{pang2024clavaddpm, solatorio2023realtabformer} have attempted to examine relationships across multiple tables within a relational database. However, such approaches remain limited in their ability to reflect inter-feature causal interactions within a single table.
We further provide a detailed summary of prior studies on tabular data generation in~\Cref{appendix:extended_related_work}.
To bridge these gaps, we introduce TabStruct, a unified evaluation framework, along with global utility, an SCM-free metric that quantifies the preservation of causal structures in tabular data.


\begin{table}[!tbp]
    \centering
    \caption{\textbf{Evaluation scope comparison between TabStruct and prior tabular generative modelling benchmarks.} TabStruct presents a comprehensive evaluation framework for tabular generative models, incorporating a wide range of evaluation dimensions, datasets, and generator categories.}
    \label{tab:literature-benchmark-overview}
    \resizebox{\textwidth}{!}{%
    \begin{tabular}{l|ccc|cc|cc|cc}
        \toprule
        \multirow{2}{*}{\textbf{Benchmark}}             & \multicolumn{3}{c}{\textbf{Conventional dimensions}}            & \multicolumn{2}{|c}{\textbf{Structural fidelity}}                                                                                                                            & \multicolumn{2}{|c|}{\textbf{Data}}            & \multicolumn{2}{c}{\textbf{Generator}}   \\
                                                        & Density Estimation                               & Privacy Preservation                              & ML Efficacy                               & SCM data & Real-world data            & \# Datasets & Contamination-free                    & \# Models    & \# Categories              \\
        \midrule
        \citet{hansen2023reimagining}      & {\color[HTML]{3A7B21} \CheckmarkBold} & {\color[HTML]{E6341C} \XSolidBrush}   & {\color[HTML]{3A7B21} \CheckmarkBold} & {\color[HTML]{E6341C} \XSolidBrush}   & {\color[HTML]{E6341C} \XSolidBrush}                              & 11         & {\color[HTML]{3A7B21} \CheckmarkBold} & 5           & 5                      \\
        Synthcity~\citep{qian2024synthcity}              & {\color[HTML]{3A7B21} \CheckmarkBold} & {\color[HTML]{3A7B21} \CheckmarkBold} & {\color[HTML]{3A7B21} \CheckmarkBold} & {\color[HTML]{E6341C} \XSolidBrush}   & {\color[HTML]{E6341C} \XSolidBrush}                              & 18         & {\color[HTML]{E6341C} \XSolidBrush}   & 6           & 4                      \\
        SynMeter~\citep{du2024systematic}                & {\color[HTML]{3A7B21} \CheckmarkBold} & {\color[HTML]{3A7B21} \CheckmarkBold} & {\color[HTML]{3A7B21} \CheckmarkBold} & {\color[HTML]{E6341C} \XSolidBrush}   & {\color[HTML]{E6341C} \XSolidBrush}                              & 12         & {\color[HTML]{E6341C} \XSolidBrush}   & 8           & 4                      \\
        CauTabBench~\citep{tu2024causality}              & {\color[HTML]{3A7B21} \CheckmarkBold} & {\color[HTML]{E6341C} \XSolidBrush}   & {\color[HTML]{E6341C} \XSolidBrush}   & {\color[HTML]{3A7B21} \CheckmarkBold} & {\color[HTML]{E6341C} \XSolidBrush}                              & 10         & {\color[HTML]{3A7B21} \CheckmarkBold} & 7           & 4                      \\
        \citet{livieris2024evaluation}              & {\color[HTML]{3A7B21} \CheckmarkBold} & {\color[HTML]{E6341C} \XSolidBrush}   & {\color[HTML]{E6341C} \XSolidBrush}   & {\color[HTML]{E6341C} \XSolidBrush} & {\color[HTML]{E6341C} \XSolidBrush}                              & 2         & {\color[HTML]{3A7B21} \CheckmarkBold} & 5           & 2                      \\
        SynthEval~\citep{lautrup2025syntheval}         & {\color[HTML]{E6341C} \XSolidBrush}   & {\color[HTML]{3A7B21} \CheckmarkBold} & {\color[HTML]{3A7B21} \CheckmarkBold} & {\color[HTML]{E6341C} \XSolidBrush}   & {\color[HTML]{E6341C} \XSolidBrush}                              & 1          & {\color[HTML]{3A7B21} \CheckmarkBold} & 5           & 3                      \\
        \citet{kapar2025what}               & {\color[HTML]{3A7B21} \CheckmarkBold} & {\color[HTML]{E6341C} \XSolidBrush}   & {\color[HTML]{3A7B21} \CheckmarkBold} & {\color[HTML]{E6341C} \XSolidBrush}   & {\color[HTML]{E6341C} \XSolidBrush}                              & 2          & {\color[HTML]{3A7B21} \CheckmarkBold} & 6           & 4                      \\
        
        \midrule
        \rowcolor{Gainsboro!60}     
        \textbf{TabStruct (Ours) }                      & {\color[HTML]{3A7B21} \CheckmarkBold} & {\color[HTML]{3A7B21} \CheckmarkBold} & {\color[HTML]{3A7B21} \CheckmarkBold} & {\color[HTML]{3A7B21} \CheckmarkBold} & {\color[HTML]{3A7B21} \CheckmarkBold}                            & \textbf{29}         & {\color[HTML]{3A7B21} \CheckmarkBold} & \textbf{13}  & \textbf{9} \\
    
        \bottomrule
    \end{tabular}%
    }
\end{table}

\vspace{-3mm}
\section{Methods}
\label{sec:method}
\vspace{-2mm}

\subsection{Problem Setup}
\vspace{-2mm}
\textbf{Dataset and tabular generator.}
Let $\mathcal{D}_{\text{full}} \coloneqq \{(\rvx^{(i)}, y^{(i)})\}_{i=1}^{N} \sim p(\rvx, y)$ represent a labelled tabular dataset with $\rvx^{(i)} \in \mathbb{R}^{D}$. We refer to the $d$-th feature (i.e., a column/variable) as $\vx_d$, and the $d$-th feature of the $i$-th sample (i.e., a cell) as $x^{(i)}_{d}$. For notational simplicity, we define $\vx_{D+1} \coloneqq \{y^{(i)}\}_{i=1}^{N}$, so that the full collection of variables, including both features and target, can be written as $\mathcal{X} \coloneqq \{\vx_1, \dots, \vx_D, \vx_{D+1}\}$. We denote the training split of $\mathcal{D}_{\text{full}}$ as the reference dataset ($\mathcal{D}_{\text{ref}}$), and test data as $\mathcal{D}_{\text{test}}$. A tabular generator is trained on $\mathcal{D}_{\text{ref}}$ and aims to generate synthetic data $\mathcal{D}_{\text{syn}} \sim p(\tilde{\rvx}, \tilde{y})$ close to $p(\rvx, y)$.
We evaluate the quality of $\mathcal{D}_{\text{ref}}$ wrt.\ all the metrics, thus providing a benchmark performance against which $\mathcal{D}_{\text{syn}}$ is compared. We refer to any dataset being assessed as ``evaluation dataset $\mathcal{D}$'', thus, both $\mathcal{D}_{\text{ref}}$ and $\mathcal{D}_{\text{syn}}$ may serve as evaluation datasets.

\textbf{Structural causal models (SCM).}
Under the assumptions of causal sufficiency, the Markov property, and faithfulness, an SCM is defined by the quadruple $M \coloneqq \langle\mathcal{X}, \mathcal{G}, \mathcal{F}, \mathcal{E}\rangle$. $\mathcal{G}$ is the causal graph that encodes the causal relationships among the variables. $\mathcal{E} \coloneqq \{\bm{\epsilon}_{j}\}_{j=1}^{D+1}$ denotes the exogenous noise, and $\mathcal{F} \coloneqq \{f_{j}\}_{j=1}^{D+1}$ is the set of structural functions. Each variable $\vx_j$ is determined by a function $f_j$ of its parents and its exogenous noise, that is, $\vx_j = f_j\left(\text{pa}(\vx_j), \bm{\epsilon}_j\right)$, where $\text{pa}(\vx_j) \subseteq \mathcal{X} \setminus \{\vx_j\}$ denotes the parent set of $\vx_j$ in the graph $\mathcal{G}$.

\textbf{Structural fidelity.}
As an empirically effective structural prior for tabular data, SCM provides a formal framework for the underlying generative processes of tabular data~\citep{hollmann2025accurate, tu2024causality}. Therefore, we define the structural fidelity of a tabular generator as the alignment between the SCMs in its synthetic data and the ground-truth causal structures. 
We further discuss the rationales behind using causal structural prior for tabular data in~\Cref{appendix:rationales}.

\subsection{Conditional Independence: Quantifying Structural Fidelity with SCM}
\label{sec:CI_scores}
\looseness-1
\textbf{Motivation.}
We begin by quantifying structural fidelity under the assumption that the ground-truth SCM is available. Following established benchmarks in causal discovery and inference~\citep{spirtes2001causation, kaddour2022causal, tu2024causality}, we evaluate structural fidelity at the level of the Markov equivalence class. 
At this level, causal structures are represented as completed partially directed acyclic graphs (CPDAGs). The SCMs of $\mathcal{D}_{\text{ref}}$ and $\mathcal{D}_{\text{syn}}$ are equivalent if they entail the same set of conditional independence (CI) statements (see~\Cref{fig:motivation}(\circled{2}~\&~\circled{3}) for an illustration).

\textbf{CI scores at various granularities.}
Following prior work~\citep{spirtes2001causation,tu2024causality}, the full set of CI statements implied by the ground-truth SCM on $\mathcal{D}_{\text{ref}}$ is defined as
\begin{equation}
    \mathcal{C}_{\text{global}} \coloneqq 
    \bigl\{(\vx_j \independent \vx_k \mid S_{j,k}) \mid  S_{j,k}\!\subseteq\!\mathcal{X}\setminus\{\vx_j,\vx_k\}\bigr\}
    \cup
    \bigl\{(\vx_j \not\independent \vx_k \mid \widehat{S}_{j,k}) \mid \widehat{S}_{j,k}\! \subsetneq\! S_{j,k}\bigr\}
\end{equation}
where $S_{j,k}$ and $\widehat{S}_{j,k}$ are the d-separation and d-connection sets for $(\vx_j,\vx_k)$, respectively.
For each CI statement, we assess whether it holds in the evaluation dataset $\mathcal{D}$ (i.e., $\mathcal{D}_{\text{ref}}$ or $\mathcal{D}_{\text{syn}}$) by conducting a CI test at the significance level $\alpha = \text{0.01}$ via
\begin{equation}
    \widehat{\mathcal{I}}_{\alpha}(\vx_j, \vx_k \mid S_{j,k}, \widehat{S}_{j,k};\mathcal{D})
    =
    \begin{cases}
        1, &\text{if the CI statement is \textit{not} rejected on $\mathcal{D}$ at level $\alpha$},\\
        0, &\text{otherwise}.
    \end{cases}
\end{equation}
To quantify structural fidelity at varying levels of granularity, we define the CI score for any subset of CI statements $\mathcal{C} \subseteq \mathcal{C}_{\text{global}}$ as:
\begin{equation}
    \text{CI}\left(\mathcal{C}, \mathcal{D}\right)
    \;=\;
    \frac{1}{|\mathcal{C}|}\;
    \sum_{\mathcal{C}}
    \mathbbm{1} \Bigl[
        \widehat{\mathcal{I}}_{\alpha}(\vx_j, \vx_k \mid S_{j,k}, \widehat{S}_{j,k};\mathcal{D})
        \;=\;
        1
    \Bigr]
    \label{eq:CI_score}
\end{equation}
where $\text{CI}\left(\mathcal{C}, \mathcal{D}\right) \in [0, 1]$ measures the fidelity of selected CI statements in $\mathcal{D}$, and $\mathbbm{1}(\cdot)$ denotes the indicator function. A higher CI score indicates that the evaluation dataset more closely aligns with the structure of the ground-truth SCM. Implementation details for the CI scores are in~\Cref{appendix:metric_design}.

\paragraph{Local structure vs.\ Global structure.}
We assess structural fidelity at two levels of granularity: local and global. 
For local structural fidelity, we define the local CI score, $\text{CI}\left(\mathcal{C}_{\text{local}}, \mathcal{D}\right)$, by considering only the CI statements that directly involve the prediction target $y$ of a given dataset and predictive task. Specifically, we compute the local CI score using~\Cref{eq:CI_score} with $\mathcal{C}_{\text{local}} = \bigl\{(\vx_j \independent \vx_{D+1} \mid S_{j,D+1}) \mid j \in [D] \bigr\} \cup \bigl\{(\vx_j \not\independent \vx_{D+1} \mid \widehat{S}_{j,D+1}) \mid j \in [D] \bigr\}$ (see~\Cref{fig:motivation}(\circled{3}) for an illustration). 
$\mathcal{C}_{\text{local}}$ highlights which features are uninformative for predicting $y$ when conditioned on the corresponding d-separation sets. Therefore, matching the local CI set indicates which features should be ignored when learning $p(y \mid \rvx)$. A higher local CI score suggests the generator faithfully captures the local structure around the target, implying higher utility for downstream predictive tasks (\Cref{sec:exp_utility_vs_structure}). 

For global structural fidelity, we define the global CI score as the CI score computed over the full set of CI statements, that is, $\text{CI}\left(\mathcal{C}_{\text{global}}, \mathcal{D}\right)$. Global CI provides a comprehensive assessment of the entire causal structure encoded in the dataset, mitigating potential bias towards any particular variable.

\textbf{Rationales for CPDAG-level evaluation.}
Prior studies~\citep{tu2024causality, spirtes2001causation} typically evaluate the causal structure alignment at three different levels: skeleton level, Markov equivalence class level, and causal graph level.
At the skeleton level, all causal directions are ignored, resulting in a loss of information about the causal relationships between features. For instance, the causal skeleton is unable to reflect encoded physical laws shown in~\Cref{fig:motivation}. Therefore, we choose not to evaluate structural fidelity at the skeleton level due to its inability to capture reliable causal relationships across variables.
At the causal graph level, structural fidelity is assessed by comparing the directed acyclic graphs (DAGs) of the reference and synthetic datasets, which requires reliable causal discovery methods as basis. However, current causal discovery methods struggle to recover accurate DAGs from tabular data~\citep{zanga2022survey, kaddour2022causal, nastl2024causal}. Grounding structural fidelity at the DAG level would introduce additional uncertainty on top of results with questionable reliability, making it even harder to draw reliable conclusions.

In contrast, CPDAG-level evaluation strikes a good balance between evaluation efficiency and validity. Unlike full DAG constructing via causal discovery, CPDAG-level evaluation does not require the orientation of all edges, making it a more tractable yet still meaningful metric of structural fidelity. This is supported by the fact that Markov equivalent SCMs serve as minimal I-MAPs~\citep{agrawal2018minimal} of the joint distribution factorisation $p(\mathcal{X}) = \prod_{j=1}^{D+1} p(\vx_{j} \mid \text{pa} \left( \vx_{j} \right) )$, and no causal directions can be further removed. In other words, the CPDAG-level evaluation can retain sufficient real-world semantics for practical use cases, such as the physical laws in~\Cref{fig:motivation}. Therefore, TabStruct evaluates structural fidelity at the CPDAG level, balancing semantic richness with computational feasibility.
More details on the rationale for CPDAG-level evaluation are provided in~\Cref{appendix:rationales}.

\subsection{Global Utility: SCM-free Metric for Global Structural Fidelity}
\label{sec:global_utility}
\textbf{Motivation.}
The CI scores introduced in \Cref{sec:CI_scores} require access to a ground-truth SCM to enumerate the CI statements $\mathcal{C}_{\text{global}}$. However, for real-world datasets, such an SCM is typically unavailable or even non-identifiable, thereby precluding direct evaluation of structural fidelity. 
Following prior work on tabular foundation models~\citep{hollmann2025accurate}, we adopt an ``SCM-inspired and real-data-validated'' paradigm to address such a limitation. Specifically, we propose global utility as an SCM-free metric for global structural fidelity.

\textbf{Utility per variable.}
Given an evaluation dataset $\mathcal{D}$, we treat each variable $\vx_j \in \mathcal{X}$ as a prediction target. An ensemble of multiple downstream predictors is trained to predict $\vx_j$ using the remaining variables $\mathcal{X} \setminus \{\vx_j\}$ as inputs, following a standard supervised learning setup. The predictive performance on $\mathcal{D}_{\text{test}}$ is denoted as $\text{Perf}_j( \mathcal{D})$, measured using \textit{balanced accuracy} for categorical variables and \textit{root mean square error (RMSE)} for numerical variables.
We define the utility of variable $\vx_j$ as the relative performance achieved on evaluation data compared to reference data:
\begin{equation}
    \text{Utility}_j\bigl(\mathcal{D}\bigr)
    \coloneqq
    \begin{cases}
    {\text{Perf}_j \bigl(\mathcal{D}_{\text{ref}}\bigr)}^{-1}{\text{Perf}_j \bigl(\mathcal{D}\bigr)},
    & \text{if } \vx_j \text{ is categorical},\\[8pt]
    {\text{Perf}_j \bigl(\mathcal{D}\bigr)}^{-1}{\text{Perf}_j \bigl(\mathcal{D}_{\text{ref}}\bigr)},
    & \text{if } \vx_j \text{ is numerical}.
    \end{cases}
    \label{eq:utility_score}
\end{equation}
Utility offers a unified perspective for interpreting downstream performance across mixed variable types: $\text{Utility}_j \geq 1$ indicates that downstream predictors trained on $\mathcal{D}$ perform on par with or better than those trained on $\mathcal{D}_{\text{ref}}$ for predicting~$\vx_j$, whereas $\text{Utility}_j < 1$ implies a loss in predictive power. To mitigate the potential bias from a specific downstream predictor, we ensemble nine different predictors with AutoGluon~\citep{agtabular}. Full technical details are in~\Cref{appendix:metric_design}. 

\textbf{Local utility.}
We define the utility of the prediction target $y$, $\text{Utility}_{D+1}(\mathcal{D})$, as local utility, which aligns with the standard metric used to assess the ML efficacy of tabular data generators.
The theoretical (\Cref{sec:CI_scores}) and empirical (\Cref{sec:exp_utility_vs_structure}) analysis showcases a strong correlation between the local CI score~($\text{CI}\left(\mathcal{C}_{\text{local}}, \mathcal{D}\right)$) and the local utility~($\text{Utility}_{D+1}(\mathcal{D})$), suggesting that local utility is an effective measurement of the local structure around target $y$.

\textbf{Global utility.}
\looseness-1
Building on the heuristics from local utility and local structure, we further examine the relationship between global utility and global structure. We define the global utility as $\text{Global Utility}(\mathcal{D}) \coloneqq \frac{1}{D+1}\sum_{j=1}^{D+1} \text{Utility}_j(\mathcal{D})$. We hypothesise that aggregating the utility across all features can be strongly correlated with the global CI score~(i.e., $\text{CI}\left(\mathcal{C}_{\text{global}}, \mathcal{D}\right)$), as global utility is grounded in the observation that a high-fidelity generator should enable accurate conditional prediction of each variable from the others -- an idea closely tied to the Markov blanket in SCMs~\citep{fu2010markov, gao2016efficient}.
Indeed, our experiments reveal a strong correlation between global CI and global utility (\Cref{sec:exp_utility_vs_structure}), supporting that global utility serves as an effective and practical metric for evaluating global structural fidelity in the absence of ground-truth SCMs.

\textbf{Bias mitigation in global utility.}
In contrast to inherently biased local utility, the proposed global utility treats all features fairly. Specifically, we consider predicting each variable associated with different tasks (e.g., binary classification, multi-class classification, regression, etc.). A change in magnitude in predictive performance can reflect different task difficulties depending on the target variable and its type~\citep{feurer2022auto, wistuba2015learning, yogatama2014efficient, grandini2020metrics}. Consequently, absolute performance scores and their variances are not directly comparable across variables, and aggregating these scores may obscure meaningful differences across tasks~\citep{grinsztajn2022tree}.
To address this, global utility follows the standard practice~\citep{feurer2022auto, grinsztajn2022tree} to aggregate normalised utility scores (\Cref{eq:utility_score}), providing a more unified perspective on performance across heterogeneous tasks (\Cref{sec:exp_utility_vs_structure} and~\Cref{appendix:extended_discussion_validity}). 

\section{Experiments}
\label{sec:exp}

We evaluate 13 tabular generators on 29 datasets by focusing on four research questions:

\begin{itemize}[topsep=0pt, leftmargin=10pt, itemsep=0pt]
    \item \textbf{Validity of Benchmark Framework (Q1,~\Cref{sec:exp_validity_benchmark}, and~\Cref{appendix:extended_discussion_framework})}: Can the proposed evaluation framework yield valid evaluation results regarding generator performance?

    \item \textbf{Validity of Global Utility (Q2,~\Cref{sec:exp_utility_vs_structure}, and~\Cref{appendix:extended_discussion_validity})}: Can global utility serve as an effective metric for structural fidelity when ground‑truth causal structures are unavailable?
    
    \item \textbf{Structural Fidelity of Generators (Q3,~\Cref{sec:exp_generator_fidelity}, and~\Cref{appendix:extended_discussion_fidelity})}: Can existing tabular generators accurately capture the data structures across both SCM and real‑world datasets?
    
    \item \textbf{Practicability of Global Utility (Q4,~\Cref{sec:exp_practicability}, and~\Cref{appendix:extended_discussion_practicability})}: Can global utility provide stable and computationally feasible evaluation results for structural fidelity?
    
\end{itemize}

\textbf{SCM datasets.}
To reduce the gap between causal structures in SCM and real-world data, we select six expert-validated SCM datasets with 7-64 features. All SCM datasets are publicly available from \texttt{bnlearn}~\citep{scutari2011bnlearn}. Full dataset descriptions are provided in~\Cref{appendix:reproducibility}.

\textbf{Real-world datasets.}
We observe that many existing generators achieve near-perfect performance on commonly used benchmark datasets~\citep{shi2024tabdiff, zhangmixed}, suggesting that these datasets offer limited discriminative power. To address this, we select 14 classification datasets from the hard TabZilla suite~\citep{mcelfresh2024neural}, containing 846-98,050 samples and 6-145 features. We further select nine challenging regression datasets, containing 345-22,784 samples and 6-82 features. Full dataset descriptions are available in~\Cref{appendix:reproducibility}.

\textbf{Benchmark generators.}
TabStruct includes 13 existing tabular data generation methods of nine different categories. In addition, we include $\mathcal{D}_{\text{ref}}$, where the reference data is used directly for evaluation. Full implementation details are in~\Cref{appendix:benchmark_generator}.

\looseness-1
\textbf{Experimental setup.}
For each dataset of $N$ samples, we perform nested cross-validation with repeated shuffle, and the details are available in~\Cref{appendix:data_processing}. Specifically, we first split the dataset into train and test sets (80\% train and 20\% test), and further split the train set into a training split ($\mathcal{D}_{\text{ref}}$) and a validation split (90\% training and 10\% validation). For classification datasets, we perform stratified splitting to preserve the class distribution. We shuffle the dataset to repeat the splitting 10 times, summing up to 10 runs per dataset.
All benchmark generators are trained on $\mathcal{D}_{\text{ref}}$, and each generator produces a synthetic dataset with $N_{\text{ref}}$ samples.
We tune the parameterised generators using Optuna~\citep{akiba2019optuna} to minimise their average validation loss across 10 repeated runs. Each generator is given at most two hours to complete a single repeat.
The reported results are averaged by default over 10 repeats. We aggregate results across all SCM or real-world datasets because the findings are consistent across classification and regression tasks (\Cref{appendix:extended_discussion_validity}). Specifically, we use the average distance to the minimum (ADTM) metric via affine renormalisation between the top-performing and worse-performing models~\citep{grinsztajn2022tree, mcelfresh2024neural, hollmann2025accurate, margeloiutabebm, jiangprotogate}. We further provide the detailed configurations (\Cref{appendix:reproducibility}) and raw results (\Cref{appendix:extended_results}).

\begin{table}[!tbp]
    \centering
    \caption{\textbf{Benchmark results of 13 tabular generators on 29 datasets.} We report the normalised mean $\pm$ std metric values across datasets. ``N/A'' denotes that a specific metric is not applicable. We highlight the {\color[HTML]{008080} \textbf{First}}, {\color[HTML]{7030A0} \textbf{Second}} and {\color[HTML]{C65911} \textbf{Third}} best performances for each metric. For visualisation, we abbreviate ``conditional independence'' as ``CI''. The results show that the Top-3 methods in Global CI and Global utility are largely consistent between SCM and real-world datasets. This alignment suggests that the selected SCM datasets represent real-world causal structure, and global utility can serve as an effective metric for global structural fidelity when ground-truth SCM is unavailable.}
    \label{tab:summary_top3}
    \vspace{-3.5mm}
    \resizebox{\textwidth}{!}{
    \begin{tabular}{l|rrrr|rr|r|rrr}
    
    \toprule
    
    \multirow{2}{*}{\textbf{Generator}} & \multicolumn{4}{c|}{\textbf{Density   Estimation}}                             & \multicolumn{2}{c|}{\textbf{Privacy Preservation}} & \multicolumn{1}{c|}{\textbf{ML Efficacy}}                              & \multicolumn{3}{c}{\textbf{Structural Fidelity}} \\
    
    & Shape $\uparrow$ & Trend $\uparrow$ & $\alpha$-precision $\uparrow$ & $\beta$-recall $\uparrow$ & DCR $\uparrow$ & $\delta$-Presence $\uparrow$ & Local utility $\uparrow$ & Local CI $\uparrow$ & Global CI $\uparrow$ & Global utility $\uparrow$ \\
    
    \midrule
    \rowcolor{Gainsboro!60}
    \multicolumn{11}{c}{\textbf{SCM datasets}} \\
    \midrule
    
    $\mathcal{D}_{\text{ref}}$ & 1.00$_{\pm\text{0.00}}$ & 1.00$_{\pm\text{0.00}}$ & 1.00$_{\pm\text{0.00}}$ & 1.00$_{\pm\text{0.00}}$ & 0.00$_{\pm\text{0.00}}$ & 0.00$_{\pm\text{0.00}}$ & 0.99$_{\pm\text{0.01}}$ & 0.89$_{\pm\text{0.10}}$ & 1.00$_{\pm\text{0.00}}$ & 0.99$_{\pm\text{0.01}}$ \\
    
    \midrule
    
    SMOTE & {\color[HTML]{008080} \textbf{0.82}}$_{\pm\text{0.09}}$ & {\color[HTML]{008080} \textbf{0.85}}$_{\pm\text{0.06}}$ & 0.60$_{\pm\text{0.17}}$ & {\color[HTML]{008080} \textbf{0.83}}$_{\pm\text{0.01}}$ & 0.21$_{\pm\text{0.09}}$ & 0.01$_{\pm\text{0.01}}$ & {\color[HTML]{008080} \textbf{0.92}}$_{\pm\text{0.07}}$ & {\color[HTML]{008080} \textbf{0.82}}$_{\pm\text{0.12}}$ & 0.30$_{\pm\text{0.11}}$ & 0.39$_{\pm\text{0.09}}$ \\
    
    BN & {\color[HTML]{7030A0} \textbf{0.80}}$_{\pm\text{0.09}}$ & {\color[HTML]{7030A0} \textbf{0.73}}$_{\pm\text{0.10}}$ & {\color[HTML]{7030A0} \textbf{0.78}}$_{\pm\text{0.10}}$ & 0.32$_{\pm\text{0.08}}$ & {\color[HTML]{C65911} \textbf{0.65}}$_{\pm\text{0.16}}$ & 0.07$_{\pm\text{0.05}}$ & 0.41$_{\pm\text{0.17}}$ & 0.23$_{\pm\text{0.12}}$ & 0.35$_{\pm\text{0.20}}$ & 0.49$_{\pm\text{0.24}}$ \\
    
    TVAE & 0.59$_{\pm\text{0.10}}$ & 0.59$_{\pm\text{0.14}}$ & 0.65$_{\pm\text{0.14}}$ & 0.36$_{\pm\text{0.06}}$ & {\color[HTML]{7030A0} \textbf{0.70}}$_{\pm\text{0.10}}$ & 0.13$_{\pm\text{0.11}}$ & 0.78$_{\pm\text{0.13}}$ & 0.50$_{\pm\text{0.21}}$ & 0.40$_{\pm\text{0.09}}$ & 0.70$_{\pm\text{0.11}}$ \\
    
    GOGGLE & 0.46$_{\pm\text{0.16}}$ & 0.50$_{\pm\text{0.13}}$ & 0.47$_{\pm\text{0.20}}$ & 0.36$_{\pm\text{0.09}}$ & 0.55$_{\pm\text{0.13}}$ & {\color[HTML]{7030A0} \textbf{0.38}}$_{\pm\text{0.19}}$ & 0.53$_{\pm\text{0.06}}$ & 0.42$_{\pm\text{0.27}}$ & 0.14$_{\pm\text{0.03}}$ & 0.24$_{\pm\text{0.08}}$ \\
    
    CTGAN & 0.46$_{\pm\text{0.14}}$ & 0.50$_{\pm\text{0.16}}$ & 0.71$_{\pm\text{0.13}}$ & 0.34$_{\pm\text{0.08}}$ & 0.52$_{\pm\text{0.11}}$ & 0.19$_{\pm\text{0.15}}$ & {\color[HTML]{7030A0} \textbf{0.80}}$_{\pm\text{0.11}}$ & 0.61$_{\pm\text{0.08}}$ & 0.08$_{\pm\text{0.04}}$ & 0.26$_{\pm\text{0.10}}$ \\
    
    NFlow & 0.31$_{\pm\text{0.15}}$ & 0.26$_{\pm\text{0.10}}$ & 0.31$_{\pm\text{0.21}}$ & 0.15$_{\pm\text{0.09}}$ & {\color[HTML]{008080} \textbf{0.73}}$_{\pm\text{0.16}}$ & {\color[HTML]{008080} \textbf{0.51}}$_{\pm\text{0.13}}$ & 0.10$_{\pm\text{0.05}}$ & 0.09$_{\pm\text{0.07}}$ & 0.09$_{\pm\text{0.07}}$ & 0.12$_{\pm\text{0.07}}$ \\
    
    ARF & {\color[HTML]{C65911} \textbf{0.75}}$_{\pm\text{0.14}}$ & {\color[HTML]{C65911} \textbf{0.71}}$_{\pm\text{0.11}}$ & {\color[HTML]{008080} \textbf{0.79}}$_{\pm\text{0.09}}$ & 0.36$_{\pm\text{0.09}}$ & 0.50$_{\pm\text{0.13}}$ & 0.09$_{\pm\text{0.07}}$ & 0.57$_{\pm\text{0.04}}$ & 0.21$_{\pm\text{0.09}}$ & 0.35$_{\pm\text{0.11}}$ & 0.68$_{\pm\text{0.11}}$ \\
    
    TabDDPM & 0.62$_{\pm\text{0.11}}$ & 0.60$_{\pm\text{0.12}}$ & 0.64$_{\pm\text{0.19}}$ & {\color[HTML]{7030A0} \textbf{0.39}}$_{\pm\text{0.09}}$ & 0.44$_{\pm\text{0.19}}$ & 0.14$_{\pm\text{0.05}}$ & 0.29$_{\pm\text{0.06}}$ & 0.17$_{\pm\text{0.08}}$ & {\color[HTML]{7030A0} \textbf{0.69}}$_{\pm\text{0.08}}$ & {\color[HTML]{008080} \textbf{0.80}}$_{\pm\text{0.05}}$ \\
    
    TabSyn & 0.50$_{\pm\text{0.16}}$ & 0.48$_{\pm\text{0.17}}$ & 0.59$_{\pm\text{0.14}}$ & 0.31$_{\pm\text{0.11}}$ & 0.45$_{\pm\text{0.14}}$ & {\color[HTML]{C65911} \textbf{0.32}}$_{\pm\text{0.21}}$ & 0.76$_{\pm\text{0.05}}$ & {\color[HTML]{7030A0} \textbf{0.70}}$_{\pm\text{0.06}}$ & {\color[HTML]{008080} \textbf{0.70}}$_{\pm\text{0.04}}$ & {\color[HTML]{7030A0} \textbf{0.76}}$_{\pm\text{0.06}}$ \\
    
    TabDiff & 0.69$_{\pm\text{0.11}}$ & 0.62$_{\pm\text{0.15}}$ & 0.75$_{\pm\text{0.09}}$ & 0.36$_{\pm\text{0.09}}$ & 0.50$_{\pm\text{0.14}}$ & 0.13$_{\pm\text{0.03}}$ & {\color[HTML]{C65911} \textbf{0.80}}$_{\pm\text{0.06}}$ & 0.58$_{\pm\text{0.14}}$ & {\color[HTML]{C65911} \textbf{0.57}}$_{\pm\text{0.15}}$ & {\color[HTML]{C65911} \textbf{0.75}}$_{\pm\text{0.07}}$ \\
    
    TabEBM & 0.67$_{\pm\text{0.12}}$ & 0.57$_{\pm\text{0.15}}$ & {\color[HTML]{C65911} \textbf{0.76}}$_{\pm\text{0.04}}$ & 0.27$_{\pm\text{0.09}}$ & 0.55$_{\pm\text{0.22}}$ & 0.14$_{\pm\text{0.06}}$ & 0.59$_{\pm\text{0.05}}$ & 0.50$_{\pm\text{0.19}}$ & 0.26$_{\pm\text{0.11}}$ & 0.30$_{\pm\text{0.08}}$ \\
    
    NRGBoost & 0.65$_{\pm\text{0.10}}$ & 0.50$_{\pm\text{0.15}}$ & 0.61$_{\pm\text{0.14}}$ & 0.26$_{\pm\text{0.07}}$ & 0.53$_{\pm\text{0.12}}$ & 0.28$_{\pm\text{0.21}}$ & 0.75$_{\pm\text{0.01}}$ & {\color[HTML]{C65911} \textbf{0.64}}$_{\pm\text{0.05}}$ & 0.11$_{\pm\text{0.05}}$ & 0.16$_{\pm\text{0.02}}$ \\
    
    GReaT & 0.62$_{\pm\text{0.09}}$ & 0.59$_{\pm\text{0.07}}$ & 0.62$_{\pm\text{0.10}}$ & {\color[HTML]{C65911} \textbf{0.38}}$_{\pm\text{0.07}}$ & 0.52$_{\pm\text{0.07}}$ & 0.18$_{\pm\text{0.05}}$ & 0.27$_{\pm\text{0.09}}$ & 0.17$_{\pm\text{0.04}}$ & 0.16$_{\pm\text{0.05}}$ & 0.25$_{\pm\text{0.08}}$ \\
    
    \midrule
    \rowcolor{Gainsboro!60}
    \multicolumn{11}{c}{\textbf{Real-world datasets}} \\
    \midrule
    
    $\mathcal{D}_{\text{ref}}$ & 1.00$_{\pm\text{0.00}}$ & 1.00$_{\pm\text{0.00}}$ & 1.00$_{\pm\text{0.00}}$ & 1.00$_{\pm\text{0.00}}$ & 0.00$_{\pm\text{0.00}}$ & 0.00$_{\pm\text{0.00}}$ & 0.96$_{\pm\text{0.06}}$ & N/A & N/A & 0.99$_{\pm\text{0.01}}$ \\
    
    \midrule
    
    SMOTE & {\color[HTML]{C65911} \textbf{0.61}}$_{\pm\text{0.13}}$ & {\color[HTML]{008080} \textbf{0.87}}$_{\pm\text{0.05}}$ & {\color[HTML]{C65911} \textbf{0.81}}$_{\pm\text{0.11}}$ & {\color[HTML]{008080} \textbf{0.77}}$_{\pm\text{0.01}}$ & 0.19$_{\pm\text{0.09}}$ & 0.02$_{\pm\text{0.02}}$ & {\color[HTML]{008080} \textbf{0.91}}$_{\pm\text{0.07}}$ & N/A & N/A & 0.41$_{\pm\text{0.04}}$ \\
    
    BN & {\color[HTML]{008080} \textbf{0.66}}$_{\pm\text{0.11}}$ & {\color[HTML]{7030A0} \textbf{0.72}}$_{\pm\text{0.09}}$ & {\color[HTML]{008080} \textbf{0.86}}$_{\pm\text{0.09}}$ & {\color[HTML]{7030A0} \textbf{0.30}}$_{\pm\text{0.04}}$ & 0.48$_{\pm\text{0.16}}$ & 0.07$_{\pm\text{0.08}}$ & 0.38$_{\pm\text{0.16}}$ & N/A & N/A & 0.44$_{\pm\text{0.25}}$ \\
    
    TVAE & 0.45$_{\pm\text{0.20}}$ & 0.50$_{\pm\text{0.14}}$ & 0.55$_{\pm\text{0.20}}$ & 0.18$_{\pm\text{0.04}}$ & {\color[HTML]{008080} \textbf{0.68}}$_{\pm\text{0.18}}$ & 0.29$_{\pm\text{0.18}}$ & 0.70$_{\pm\text{0.06}}$ & N/A & N/A & 0.53$_{\pm\text{0.13}}$ \\
    
    GOGGLE & 0.41$_{\pm\text{0.15}}$ & 0.47$_{\pm\text{0.14}}$ & 0.57$_{\pm\text{0.16}}$ & 0.26$_{\pm\text{0.07}}$ & 0.50$_{\pm\text{0.11}}$ & {\color[HTML]{7030A0} \textbf{0.35}}$_{\pm\text{0.18}}$ & 0.46$_{\pm\text{0.04}}$ & N/A & N/A & 0.21$_{\pm\text{0.06}}$ \\
    
    CTGAN & 0.29$_{\pm\text{0.18}}$ & 0.53$_{\pm\text{0.14}}$ & 0.66$_{\pm\text{0.21}}$ & 0.11$_{\pm\text{0.05}}$ & 0.51$_{\pm\text{0.13}}$ & {\color[HTML]{C65911} \textbf{0.30}}$_{\pm\text{0.24}}$ & 0.70$_{\pm\text{0.06}}$ & N/A & N/A & 0.13$_{\pm\text{0.06}}$ \\
    
    NFlow & 0.38$_{\pm\text{0.19}}$ & 0.28$_{\pm\text{0.16}}$ & 0.52$_{\pm\text{0.15}}$ & 0.07$_{\pm\text{0.04}}$ & {\color[HTML]{7030A0} \textbf{0.64}}$_{\pm\text{0.14}}$ & {\color[HTML]{008080} \textbf{0.42}}$_{\pm\text{0.25}}$ & 0.10$_{\pm\text{0.06}}$ & N/A & N/A & 0.14$_{\pm\text{0.12}}$ \\
    
    ARF & {\color[HTML]{7030A0} \textbf{0.61}}$_{\pm\text{0.11}}$ & 0.58$_{\pm\text{0.12}}$ & {\color[HTML]{7030A0} \textbf{0.83}}$_{\pm\text{0.10}}$ & 0.21$_{\pm\text{0.04}}$ & 0.48$_{\pm\text{0.14}}$ & 0.05$_{\pm\text{0.04}}$ & 0.54$_{\pm\text{0.07}}$ & N/A & N/A & 0.56$_{\pm\text{0.12}}$ \\
    
    TabDDPM & 0.43$_{\pm\text{0.16}}$ & 0.49$_{\pm\text{0.18}}$ & 0.54$_{\pm\text{0.22}}$ & 0.26$_{\pm\text{0.09}}$ & 0.42$_{\pm\text{0.19}}$ & 0.27$_{\pm\text{0.18}}$ & 0.27$_{\pm\text{0.06}}$ & N/A & N/A & {\color[HTML]{C65911} \textbf{0.72}}$_{\pm\text{0.08}}$ \\
    
    TabSyn & 0.44$_{\pm\text{0.14}}$ & 0.51$_{\pm\text{0.16}}$ & 0.62$_{\pm\text{0.18}}$ & 0.24$_{\pm\text{0.08}}$ & 0.51$_{\pm\text{0.12}}$ & 0.24$_{\pm\text{0.14}}$ & {\color[HTML]{C65911} \textbf{0.76}}$_{\pm\text{0.08}}$ & N/A & N/A & {\color[HTML]{008080} \textbf{0.73}}$_{\pm\text{0.07}}$ \\
    
    TabDiff & 0.54$_{\pm\text{0.15}}$ & 0.52$_{\pm\text{0.16}}$ & 0.69$_{\pm\text{0.12}}$ & 0.22$_{\pm\text{0.07}}$ & 0.57$_{\pm\text{0.15}}$ & 0.20$_{\pm\text{0.13}}$ & {\color[HTML]{7030A0} \textbf{0.78}}$_{\pm\text{0.03}}$ & N/A & N/A & {\color[HTML]{7030A0} \textbf{0.73}}$_{\pm\text{0.07}}$ \\
    
    TabEBM & 0.59$_{\pm\text{0.15}}$ & {\color[HTML]{C65911} \textbf{0.65}}$_{\pm\text{0.08}}$ & 0.79$_{\pm\text{0.04}}$ & {\color[HTML]{C65911} \textbf{0.30}}$_{\pm\text{0.10}}$ & {\color[HTML]{C65911} \textbf{0.58}}$_{\pm\text{0.16}}$ & 0.14$_{\pm\text{0.03}}$ & 0.63$_{\pm\text{0.11}}$ & N/A & N/A & 0.35$_{\pm\text{0.11}}$ \\
    
    NRGBoost & 0.54$_{\pm\text{0.12}}$ & 0.49$_{\pm\text{0.13}}$ & 0.62$_{\pm\text{0.16}}$ & 0.20$_{\pm\text{0.07}}$ & 0.51$_{\pm\text{0.15}}$ & 0.22$_{\pm\text{0.13}}$ & 0.74$_{\pm\text{0.05}}$ & N/A & N/A & 0.16$_{\pm\text{0.05}}$ \\
    
    GReaT & 0.47$_{\pm\text{0.10}}$ & 0.49$_{\pm\text{0.13}}$ & 0.57$_{\pm\text{0.14}}$ & 0.26$_{\pm\text{0.08}}$ & 0.52$_{\pm\text{0.11}}$ & 0.27$_{\pm\text{0.15}}$ & 0.23$_{\pm\text{0.07}}$ & N/A & N/A & 0.20$_{\pm\text{0.06}}$ \\
    
    \bottomrule
    
    \end{tabular}
    }
\end{table}

\vspace{-2mm}
\subsection{Validity of Benchmark Framework (Q1)}
\label{sec:exp_validity_benchmark}
\vspace{-1mm}

\textbf{The benchmark results effectively evaluate data quality.}
\Cref{tab:summary_top3} demonstrates that all metrics effectively distinguish between high- and low-quality data. Specifically, except for privacy-related metrics, the reference data ($\mathcal{D}_{\text{ref}}$) consistently achieves the highest scores. This is expected, as $\mathcal{D}_{\text{ref}}$ is the ground truth and should score highly on metrics of density estimation, ML efficacy, and structural fidelity. In contrast, privacy metrics reward greater differences from the ground truth to indicate stronger privacy preservation. Since $\mathcal{D}_{\text{ref}}$ is identical to the ground truth, it naturally scores poorly for privacy.
These results show that the selected metrics provide appropriate evaluations for data quality. Therefore, we consider the evaluation results to be valid and meaningful for analysis.

\looseness-1
\textbf{Structural fidelity is complementary to conventional evaluation dimensions, rather than interchangeable.}
On SCM datasets, \Cref{fig:scm_metric_corr} (left) shows that none of the existing evaluation metrics exhibits a strong correlation with global CI. Notably, SMOTE and BN tend to outperform other models by a clear margin in density estimation. However, their performance degrades greatly when it comes to capturing the global structure of tabular data, as reflected by global CI, consistent with our motivating example in~\Cref{fig:motivation}.
This discrepancy reveals the limitations of conventional evaluation dimensions and underscores the need to incorporate structural fidelity for inter-feature causal structures.

\subsection{Validity of Global Utility (Q2)}
\label{sec:exp_utility_vs_structure}

\begin{figure}[!t]
    \centering
    \includegraphics[width=\textwidth]{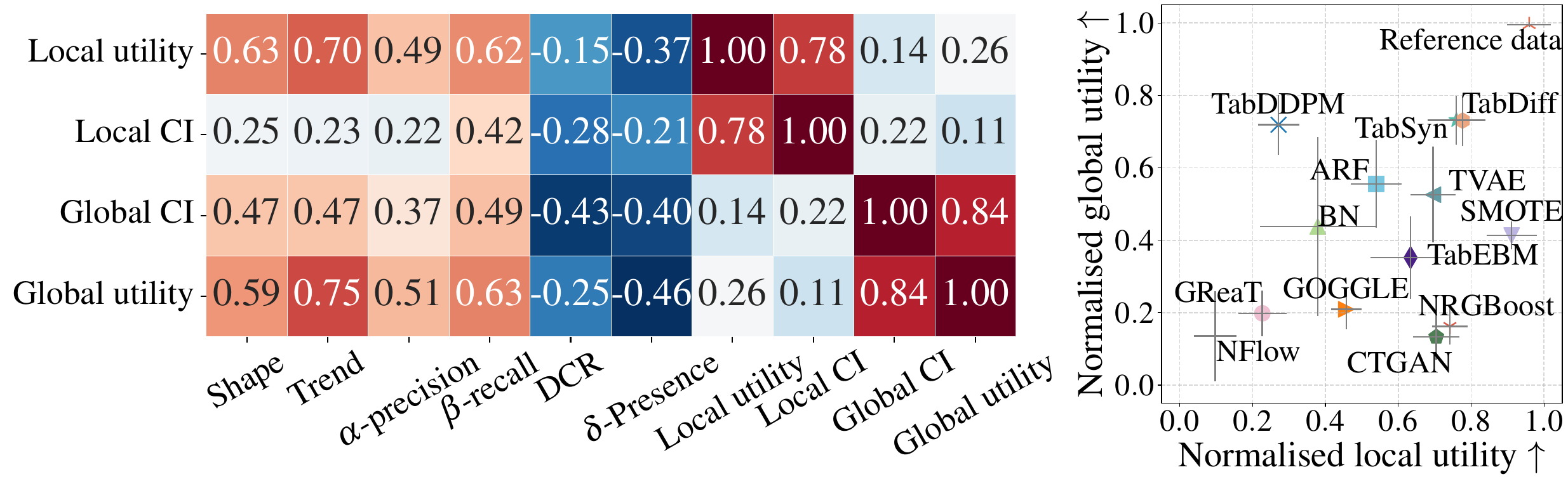}
    \caption{\textbf{Left:} Spearman's rank correlation heatmap based on metric values on six SCM datasets. Global utility correlates strongly with global CI, suggesting that global utility can effectively assess global structural fidelity without resorting to SCMs. \textbf{Right:} Mean normalised local utility vs.\ mean normalised global utility on 23 real-world datasets. SMOTE prioritises local utility, whereas TabDiff and TabSyn generally achieve a balanced preservation of both global and local data structures.}
\label{fig:scm_metric_corr}
\end{figure}

\looseness-1
\textbf{Global utility serves as an effective metric for global structural fidelity.}
\Cref{tab:summary_top3} and~\Cref{fig:scm_metric_corr} (left) demonstrate a strong monotonic correlation between global utility and global CI scores (\mbox{$r_s=0.84, \text{p}<0.001$}). 
To ensure the generalisability of global utility, we extend our evaluation scope, incorporating more complex SCM datasets~(\Cref{appendix:extended_discussion_validity}), a wider range of existing metrics~(\Cref{appendix:extended_discussion_validity}), and additional downstream tasks~(\Cref{appendix:extended_discussion_practicability}). Across all settings, global utility consistently exhibits a substantially stronger correlation with global CI than any other metric.
\Cref{appendix:extended_discussion_validity} further shows that global utility more closely aligns with global CI in the induced generator rankings. 
We would like to emphasise that the high correlation between global CI and global utility is an empirical finding rather than a formal proof of a connection between the two metrics. This observation primarily aims to offer users actionable and empirically grounded insights into tabular data generation.
The strong correlation and consistent generator ranking suggest that global utility offers a robust, SCM-free approach for assessing global structural fidelity.

\textbf{Local utility is not always the golden standard, due to its bias towards the local structure.}
We further examine the correlation between local utility and local CI, which only considers the local structure associated with the prediction target. As shown in~\Cref{fig:scm_metric_corr} (left), local utility exhibits a strong correlation with local CI ($r_s=0.78, \text{p}<0.001$), but a much weaker correlation with global CI ($r_s=0.14, \text{p}<0.001$). The results indicate that local utility may overlook the holistic data structure, while global utility provides a more comprehensive evaluation of structural fidelity.

\subsection{Structural Fidelity of Generators (Q3)}
\label{sec:exp_generator_fidelity}

\looseness-1
\textbf{Structure learning methods struggle with tabular data generation.}
One surprising finding is that BN and GOGGLE do not demonstrate strong performance in structural fidelity, despite their inductive bias towards learning tabular data structures. This observation aligns with prior work~\citep{tu2024causality, zeng2022causal}, which highlights that current causal discovery algorithms often struggle when the number of features exceeds 10 -- our benchmark datasets have features from 6 up to 145. Furthermore, GOGGLE exhibits notable performance degradation when prior knowledge about the data structure is missing~\citep{liu2023goggle}. The results underscore the limitations of existing causal discovery methods in recovering precise causal structures from real-world data, further justifying our evaluation at the CPDAG level. 

\textbf{Diffusion models generally capture the global structure well.}
As reported in~\Cref{tab:summary_top3} and~\Cref{fig:scm_metric_corr} (right), diffusion-based models consistently achieve the highest scores in global structural fidelity: the Top-3 methods are TabDDPM, TabSyn, and TabDiff across both SCM and real-world datasets. 
We attribute their strong performance to the inherent learning principle of diffusion models for learning permutation-invariant conditional distributions of each feature. 
At the training stage, since noise is added independently to each feature, the diffusion network is optimised at every denoising step to reconstruct each feature simultaneously by conditioning on others. For instance, TabDDPM and TabDiff implement this principle within each feature type, and TabSyn applies it across all features. 
Moreover, diffusion models impose no ordering of features. This results in efficient computation (\Cref{fig:computation_time}) and permutation-invariant conditional distributions, a property that aligns naturally with the structure of tabular data. These theoretical properties align with the conditional independence analysis in~\Cref{sec:CI_scores}, thus confirming that diffusion models are capable of capturing global structure.

\looseness-1
\textbf{Language models remain limited in learning tabular data structure.}
\Cref{tab:summary_top3} shows that the autoregressive model GReaT, even with the help of large language models, fails to outperform even the simple baselines like SMOTE and TVAE.
Although token-wise likelihood training is a well-established approach for sequential modalities like text and time series, its underlying assumptions misalign with the permutation-invariant nature of tabular data. An autoregressive generator needs to linearise the feature set and then factorise the joint distribution as $\prod_{j=1}^{d} p(\rvx_{\pi(j)} \mid \rvx_{\pi(<j)})$, where $\pi$ denotes a chosen ordering of features. Any fixed ordering $\pi$ can introduce directional bias. For instance, the bias could harm the estimation of $p(\vx_{j} \mid \mathcal{X} \setminus \{\vx_j\})$ when $j$ appears early in the sequence.
While GReaT attempts to mitigate this issue by randomising $\pi$ when fine-tuning large language models, randomising $\pi$ does not resolve the fundamental misalignment and can even constrain the performance of autoregressive tabular generators (\Cref{appendix:extended_discussion_fidelity}).

\subsection{Practicability of Global Utility (Q4)}
\label{sec:exp_practicability}

\looseness-1
\textbf{Global utility is robust and stable.}
\Cref{appendix:global_utility_computation} and~\Cref{appendix:extended_discussion_practicability} show that global utility yields stable generator rankings across both nine tuned predictors (``Full-tuned'') and three untuned ones (``Tiny-default''). In contrast, local utility necessitates nine tuned predictors (``Full-tuned'') for reliable results.
These findings align with the conceptual design of the utility metrics. 
We note that local utility focuses mainly on the predictive performance of a single target variable, making it susceptible to feature-specific bias, which results in unstable generator rankings across different predictor configurations. In contrast, global utility aggregates performance across all variables, thereby mitigating feature-specific effects and enhancing robustness.

\begin{figure}[!tp]
    \centering
    \includegraphics[width=0.93\textwidth]{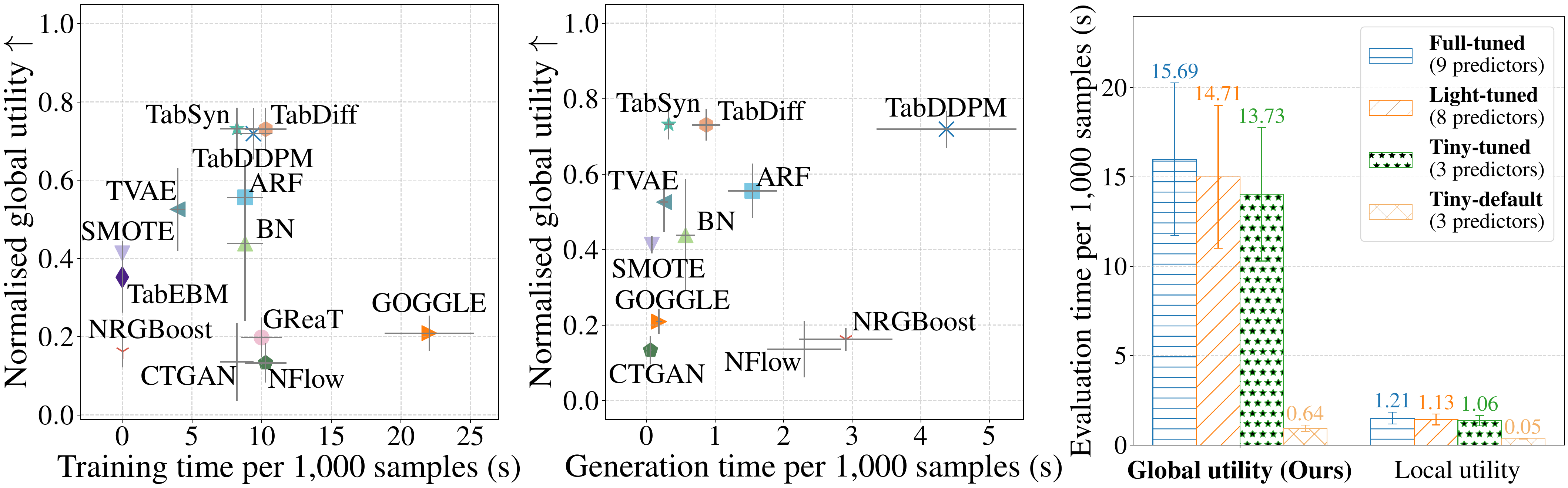}
    \caption{\textbf{Computation efficiency on 23 real-world datasets.} \textbf{Left:} Median training time per 1,000 samples vs.\ mean normalised global utility. \textbf{Middle:} Median generation time per 1,000 samples vs.\ mean normalised global utility. We exclude the outliers (TabEBM and GReaT) due to their long generation time (over 30s). \textbf{Right:} Median evaluation time. Because global utility yields stable generator rankings across downstream predictors (\Cref{appendix:extended_discussion_practicability}), computing global utility can be highly efficient with only a small ensemble of predictors (i.e., Tiny-default).}
\label{fig:computation_time}
\end{figure}

\textbf{Global utility provides efficient evaluations of structural fidelity.}
In practice, we are often interested in identifying the most promising model before fine-tuning it for optimal performance. Global utility supports this by reducing both the tuning burden and the dependency on the number of predictors, while still yielding stable and informative rankings. As illustrated in \Cref{fig:computation_time} (right), computing global utility with ``Tiny-default'' takes only 0.64s per 1000 samples, while local utility requires nearly double the time (``Full-tuned'' with 1.21s) for comparable reliability.

\textbf{Limitations and future work.} 
While our proposed global utility is a robust and effective metric for assessing global structural fidelity, it is an empirical measurement of the likely SCMs behind the data at hand. However, developing a theoretically provable structural fidelity metric for real-world tabular data is highly challenging, as ground-truth causal structures are rarely available, even precluding the possibility of theoretical validation. This is in line with several open challenges in the field -- particularly the lack of causal discovery methods that can reliably infer the governing SCMs of real-world tabular datasets~\citep{kaddour2022causal, tu2024causality, glymour2019review, nastl2024causal}. 
Despite substantial research efforts, recent work~\citep{nastl2024causal} shows that even state-of-the-art causal discovery methods often perform poorly on real-world data and may mislead users. Therefore, we propose global utility primarily as an empirical lens for evaluating tabular data structures. Bridging the gap between theoretical assumptions and real-world causal structures will require advances in causal modelling. As TabStruct library is freely available, its development will be an ongoing, community-driven endeavour. Therefore, TabStruct will continue to evolve with advances in causal modelling. We believe that the open-source nature of TabStruct will help drive progress in theoretical foundations for real-world tabular data challenges.
More discussion on future work is in~\cref{appendix:extended_discussion_guidance} and~\Cref{appendix:extended_discussion_future}. 

\section{Conclusion}

We present TabStruct, a principled benchmark for tabular data generators along with both structural fidelity and conventional dimensions. To address the challenge of assessing structural fidelity in the absence of ground-truth SCMs, we introduce global utility -- a novel, SCM-free metric that enables unbiased and holistic evaluation for tabular data structure. 

In our large-scale study of 13 generators across 29 datasets, we find that existing evaluation methods often favour models that capture local causal interactions while neglecting global structure. Our results show that the four evaluation dimensions are complementary, offering practical guidance for selecting suitable generators across diverse applications. We further observe that diffusion models, due to their permutation-invariant generation process, offer valuable insights into the fundamental representation learning of tabular data.
TabStruct is an ongoing effort. As such, it will continue to evolve with additional datasets, generators, and evaluation metrics -- both through our engagement and contributions from the community. We envision that the open-source nature of TabStruct will help drive progress in high-fidelity tabular generative modelling.

\section*{Ethics Statement}
This paper proposes integrating structural fidelity as a core evaluation dimension alongside conventional metrics for assessing tabular data generators. Specifically, we introduce global utility, a novel metric that evaluates the structural fidelity of synthetic tabular data without requiring access to the ground-truth causal structures. 
Furthermore, we present TabStruct, a comprehensive benchmark for tabular data generation that spans a wide evaluation scope -- comprising 13 generators from nine distinct categories, evaluated on 29 datasets. Our benchmark results highlight that structural fidelity is an important yet previously underexplored evaluation dimension. It effectively captures whether generated data preserves the underlying causal structures present in real-world tabular datasets, serving as a valuable complement to existing evaluation dimensions.

This is particularly critical for tabular modalities, where visual inspection of data authenticity is not feasible, unlike in text or image domains~\citep{van2024tabular, zhao2023survey}. 
By providing a unified benchmark that incorporates both conventional metrics and structural fidelity, TabStruct has the potential to foster more reliable and transparent development of generative models. This can benefit multiple domains that rely on tabular data, such as healthcare~\citep{jiangprotogate, bespalov2016failed, morford2011preclinical} and scientific research~\citep{margeloiutabebm}, where understanding the structural fidelity of generated data is crucial.

The impact of our work extends to enabling broader machine learning applications in data-scarce domains. For instance, it can facilitate robust data analysis in clinical contexts where data collection is limited~\citep{margeloiutabebm, chawla2002smote, mclachlan2018aten}. Enhancing the fidelity of synthetic data may promote the adoption of more advanced machine learning approaches. TabStruct could further facilitate safer data sharing in privacy-sensitive contexts~\citep{jordon2018pate, hu2024sok, stoian2025survey, alami2020artificial, ciecierski2022artificial}, support reproducible research through synthetic benchmarks, and broaden access to machine learning capabilities in low-resource or data-scarce scenarios.

\section*{Reproducibility Statement}
Our study is conducted entirely within a reproducible setting. As detailed in~\Cref{appendix:reproducibility}, all benchmark datasets are publicly available and widely adopted in the machine learning literature~\citep{mcelfresh2024neural, scutari2011bnlearn}. We do not use, include, or release any newly collected or proprietary data.
In addition, the employed tabular generative models and benchmark metrics are not tailored to any specific demographic or domain-sensitive dataset. Full implementation details are available in~\Cref{appendix:reproducibility} and the associated codebase (\url{https://github.com/SilenceX12138/TabStruct}).
Furthermode, we release TabStruct as an open-source library to support transparency, reproducibility, and further community-driven development. We welcome community contributions that prioritise safety, fairness, and inclusivity in the future evolution of the benchmark.

\section*{Acknowledgements}

The authors would like to express their gratitude to Prof.~Carl Henrik Ek for insightful discussions on structure learning, and to Prof.~Ferenc Huszár and Dr.~Ruibo Tu for their enlightening perspectives on causal machine learning, and to Dr.~Andrei Margeloiu for his thoughtful feedback during the early stages of the project.
XJ acknowledges the generous support of the Google PhD Fellowship.
MJ and NS acknowledge the support of the U.S. Army Medical Research and Development Command of the Department of Defense; through the FY22 Breast Cancer Research Program of the Congressionally Directed Medical Research Programs, Clinical Research Extension Award GRANT13769713. Opinions, interpretations, conclusions, and recommendations are those of the authors and are not necessarily endorsed by the Department of Defense.

\bibliography{reference}
\bibliographystyle{style/iclr_2026/iclr2026_conference}


\input{paper_3_appendix_iclr_camera}

\end{document}